\documentclass[conference]{IEEEtran}
\usepackage{cite}
\usepackage{amsmath,amssymb,amsfonts}
\usepackage{algorithmic}
\usepackage{graphicx}
\usepackage{textcomp}
\usepackage{xcolor}
\usepackage{booktabs}
\usepackage{array}
\usepackage{url}
\usepackage{hyperref}

\hypersetup{
    colorlinks=true,
    linkcolor=blue,
    citecolor=blue,
    urlcolor=blue
}

\begin{document}

\title{Fault Detection and Bit-Error Rate Diagnostics in Optical Wireless Links Using Edge-AI}

\author{
\IEEEauthorblockN{Sarthak Bhattarai, Sheshadri Shikhar Bandyopadhyay, Suprabh Dhakal, Jitendra Kumar Manandhar}
\IEEEauthorblockA{\textit{Department of Electronics and Computer Engineering} \\
\textit{Pulchowk Campus, Institute of Engineering, Tribhuvan University}\\
Lalitpur, Nepal \\
Email: \{080bei038.sarthak, 080bei040.sheshadri, 080bei043.suprabh\}@pcampus.edu.np, mejiten@ioe.edu.np}
}

\maketitle

\begin{abstract}
Optical Wireless Communication (OWC) systems provide high-bandwidth, spatially secure, and RF-interference-free digital transmission. However, direct free-space optical links suffer from extreme sensitivity to line-of-sight disruptions, optical attenuation, high ambient light saturation, and transmitter-receiver misalignment. Traditional channel diagnostic frameworks rely on external computing nodes or cloud-based processing, which introduces severe diagnostic latency and reduces edge node autonomy. This paper presents an autonomous, self-diagnosing optical wireless receiver node that integrates physical-layer signal telemetry extraction with an on-device TinyML classification engine. Built around an ESP32-S3 microcontroller, the receiver recovers $1000~\text{bit/s}$ On-Off Keying (OOK) digital payloads using a 5-sample majority voting mechanism while concurrently extracting real-time channel features: Bit Error Rate (BER), received signal voltage ($V_{\text{signal}}$), Signal-to-Noise Ratio ($\text{SNR}_{\text{dB}}$), and ambient light baseline ($L_{\text{ambient}}$). Evaluated against a Random Forest baseline ($77.22\%$ accuracy) and a floating-point neural network ($78.48\%$ accuracy), a post-training INT8 quantized TensorFlow Lite Micro model occupying only $3.32~\text{kB}$ of SRAM runs directly on the receiver firmware. The quantized model achieves $74.68\%$ overall accuracy across four distinct operational conditions: normal transmission, optical filter attenuation, ambient light interference, and severe obstruction/misalignment. On-device deployment tests confirm real-time, deterministic fault classification, establishing a practical approach for self-aware optical IoT nodes.
\end{abstract}


\section{Introduction}
Optical Wireless Communication (OWC), encompassing Visible Light Communication (VLC) and Free-Space Optics (FSO), uses optical spectrum frequencies to transmit digital data \cite{khan2018vlc}. Offering immunity to radio-frequency (RF) interference, high spatial reuse, and strict spatial confinement, OWC serves as an ideal supplement to congested indoor RF networks \cite{sejan2023mimo}. However, narrow-beam directed optical links depend on strict Line-of-Sight (LOS) alignment between the optical emitter and photodetector \cite{purnita2021ml}. Physical misalignments, transient human or structural blockages, optical signal attenuation, and high background illumination degrade communication quality or cause silent channel failure \cite{salman2025ml}.

\begin{figure}[htbp]
\centering
\includegraphics[width=\columnwidth]{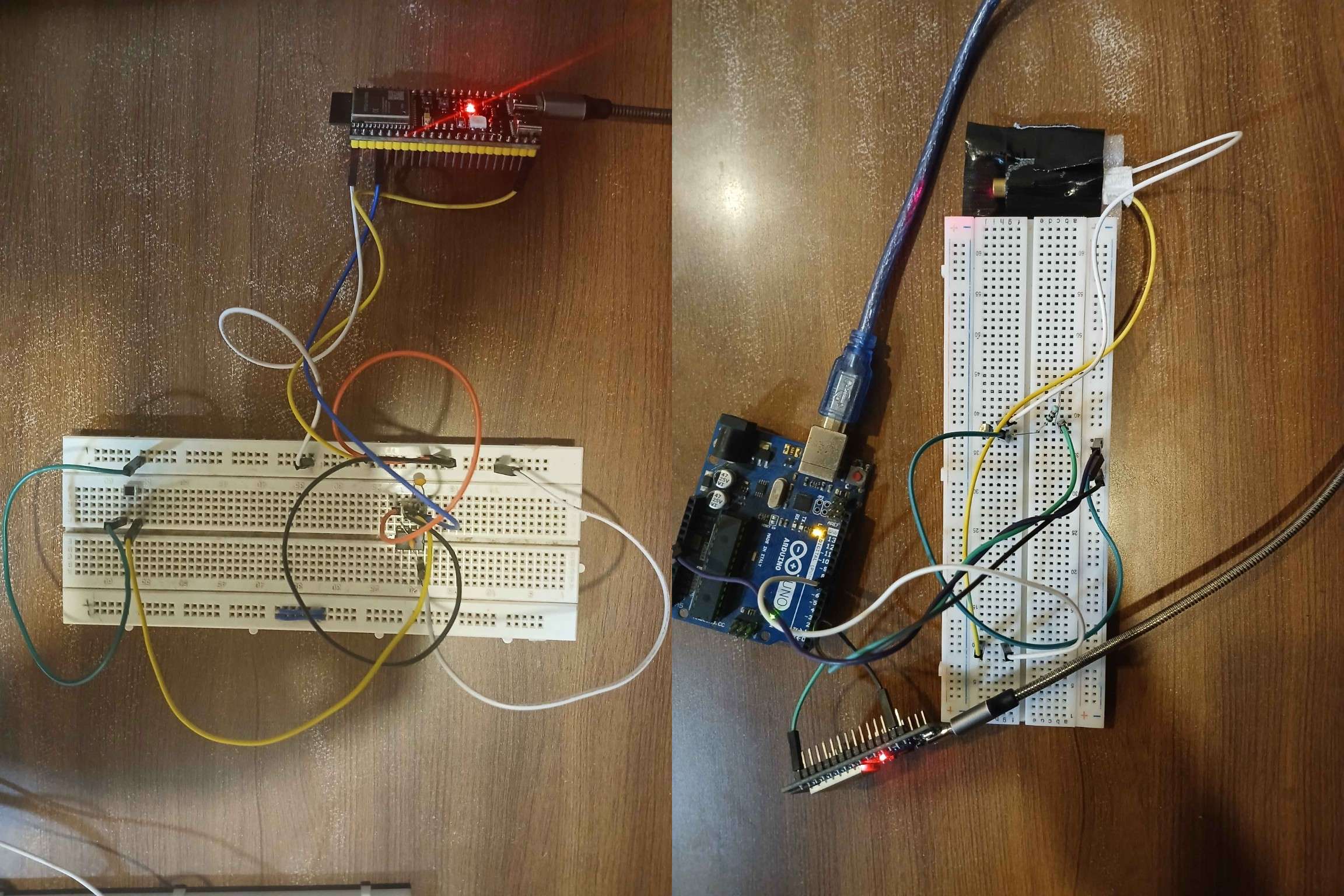}
\caption{Complete physical hardware setup of the autonomous optical wireless communication testbed.}
\label{fig:hardware_setup}
\end{figure}

In conventional OWC architectures, physical-layer health parameters (e.g., Bit Error Rate or Received Signal Strength) remain unmonitored or offload to external host computers or cloud infrastructure \cite{purnita2021ml}. Offloading introduces network transmission overhead, processing latency, and total dependency on external infrastructure. Figure~\ref{fig:hardware_setup} demonstrates our physical experimental testbed designed to eliminate external computing dependencies.

\begin{figure}[htbp]
\centering
\includegraphics[width=\columnwidth]{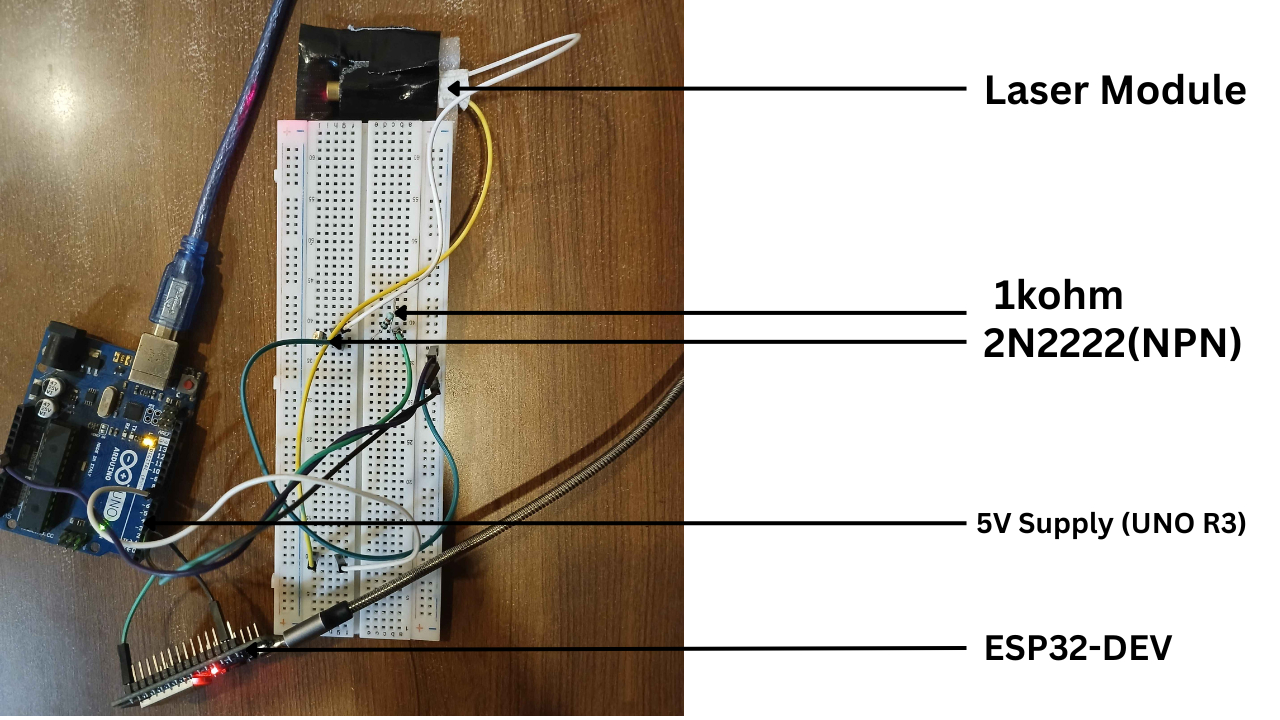}
\caption{Transmitter circuit schematic showing GPIO interfacing, transistor switching logic, and laser driver circuit.}
\label{fig:circuit_tx}
\end{figure}

\begin{figure}[htbp]
\centering
\includegraphics[width=\columnwidth]{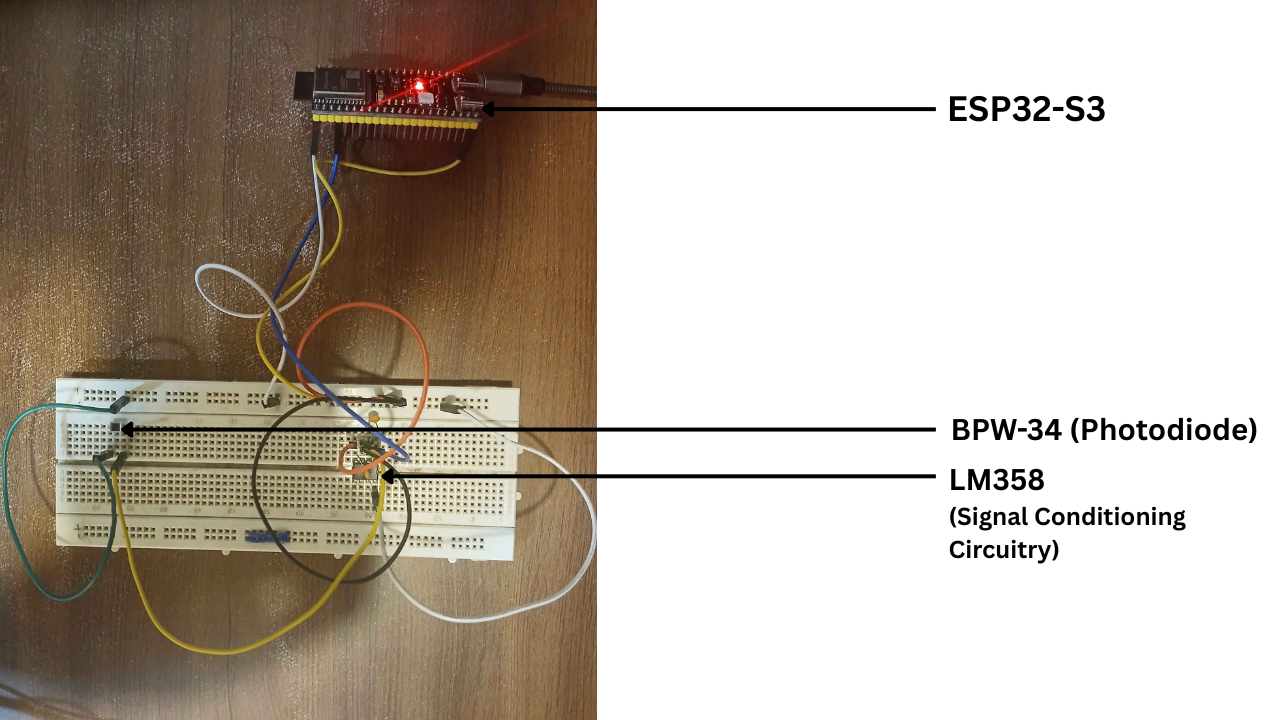}
\caption{Receiver circuit schematic depicting BPW34 PIN photodiode photodetector, LM358 op-amp conditioning, and ESP32-S3 ADC acquisition.}
\label{fig:circuit_rx}
\end{figure}

To resolve these challenges, we present a fully integrated, edge-based optical communication and diagnostic system. Operating entirely on an ultra-low-power ESP32-S3 microcontroller, the receiver recovers digital data payloads while extracting physical-layer telemetry parameters \cite{warden2019tinyml}. The physical circuit schematics of the transmitter and receiver front-ends are shown in Fig.~\ref{fig:circuit_tx} and Fig.~\ref{fig:circuit_rx}, respectively. A lightweight neural network quantized to 8-bit integers (INT8) executes directly on the receiver firmware using TensorFlow Lite Micro, enabling real-time, autonomous fault classification without external computation.

\section{Literature Review}
Research in Optical Wireless Communication traditionally focuses on modulation efficiency, channel capacity, and transmission throughput. Khan \cite{khan2018vlc} provides an extensive architectural review of VLC systems, highlighting Line-of-Sight constraints, ambient light interference, and link vulnerability as primary operational challenges.

Machine learning applications in optical links show significant performance improvements in channel estimation and adaptive modulation. Purnita and Mondal \cite{purnita2021ml} apply supervised machine learning to DCO-OFDM LiFi links to analyze optical communication parameters and enhance channel capacity under varying ambient noise. Similarly, Sejan et al. \cite{sejan2023mimo} survey MIMO-VLC links and emphasize the necessity of intelligent adaptation mechanisms to manage spatial receiver alignment errors and optical power fluctuations.

Adaptive transmission control forms another significant research direction. Salman et al. \cite{salman2025ml} develop a machine learning framework that predicts DC bias and adaptive transmission parameters for indoor OFDM LiFi networks, maintaining target signal quality under variable channel conditions. In FSO links, Nistazakis, Tsiftsis, and Tombras \cite{nistazakis2009fso} evaluate physical channel impairments and atmospheric turbulence, demonstrating that signal voltage degradation and SNR drops serve as primary indicators of physical link quality.

Despite these advancements, conventional channel monitoring architectures offload computation to centralized processing units. You et al. \cite{you2022edge} demonstrate that centralized computation offloading introduces diagnostic transmission latency and processing energy consumption overhead, severely limiting edge node autonomy.

The emergence of Tiny Machine Learning (TinyML) shifts execution directly to low-power microcontrollers. Warden and Situnayake \cite{warden2019tinyml} establish that quantized neural networks execute on resource-constrained hardware with minimal memory footprints. David et al. \cite{david2021tensorflow} demonstrate TensorFlow Lite Micro deployment on ARM Cortex and ESP32 targets, achieving sub-millisecond inference times for local sensor classification.

In physical-layer monitoring, Dutta and Roy \cite{dutta2023edge} investigate real-time signal diagnostics and fault detection in embedded communication nodes. However, past studies focus primarily on RF channels or high-power computing units.

A research gap remains in combining physical-layer optical metrics ($V_{\text{signal}}$, BER, SNR, $L_{\text{ambient}}$) with quantized TinyML models deployed directly on microcontroller receivers. Our work fills this gap by building an autonomous ESP32-S3 optical receiver that recovers data payloads and classifies four physical channel conditions locally.

\section{System Architecture and Mathematical Modeling}
\begin{figure}[htbp]
\centering
\includegraphics[width=\columnwidth]{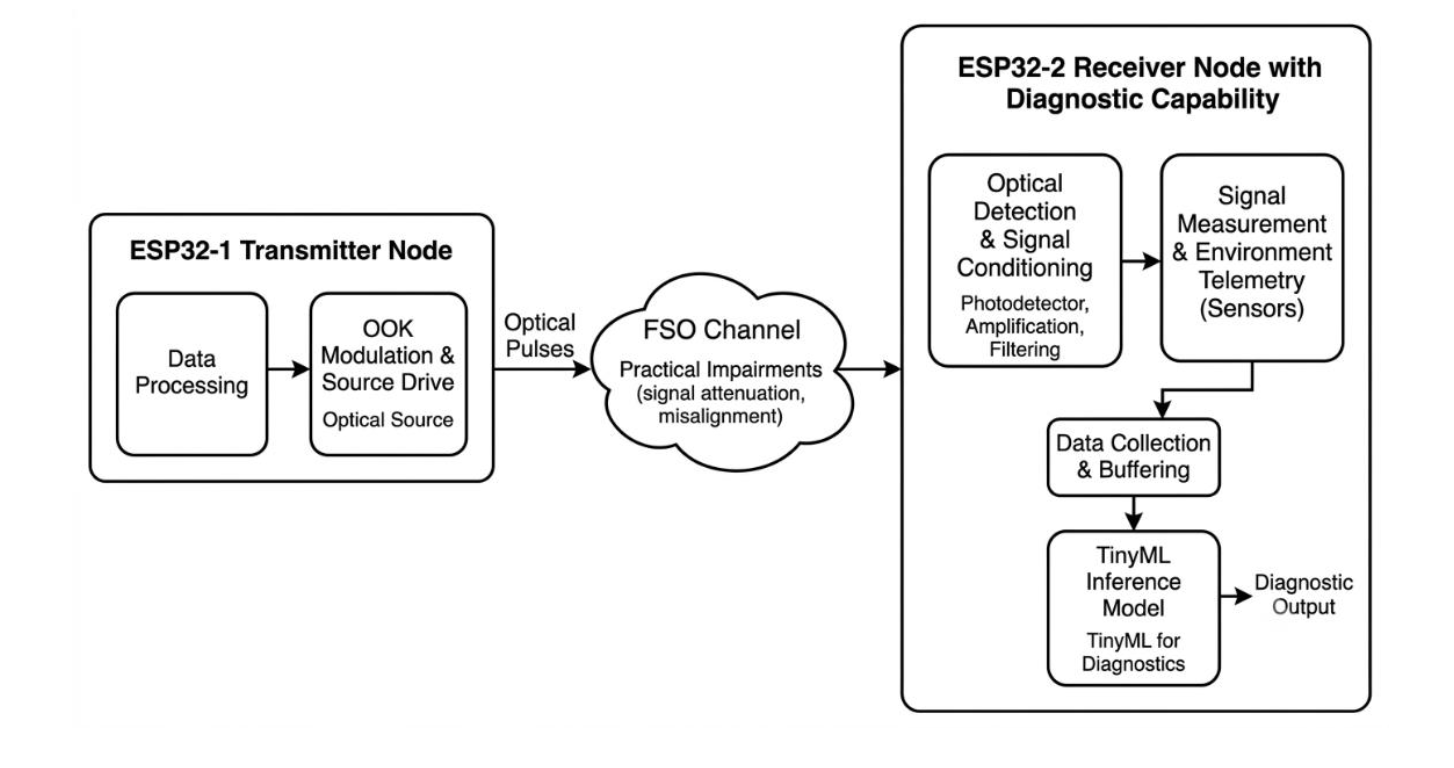}
\caption{Overall framework of the autonomous OWC testbed illustrating transmitter, optical channel, and receiver-side TinyML inference engine.}
\label{fig:sys_arch}
\end{figure}

The system architecture consists of three principal modules: an optical transmitter node, a free-space optical propagation channel, and an edge-diagnostic optical receiver node, as illustrated in Fig.~\ref{fig:sys_arch}.

\subsection{Optical Data Representation and Modulation}
The system employs On-Off Keying (OOK) modulation with Non-Return-to-Zero Level (NRZ-L) line coding. The relationship between bit rate $R_b$ and bit duration $T_b$ follows:
\begin{equation}
    R_b = \frac{1}{T_b}
\end{equation}
For $R_b = 1000~\text{bit/s}$, each bit occupies $T_b = 1~\text{ms}$. A logical '1' corresponds to laser activation for the full bit duration, whereas a logical '0' corresponds to laser deactivation.

\subsection{Physical-Layer Telemetry Extraction}
During each transmission sequence, the receiver extracts four diagnostic features forming a feature vector $\mathbf{x} = [V_{\text{signal}}, \text{BER}, \text{SNR}_{\text{dB}}, L_{\text{ambient}}]^T$:

\subsubsection{Bit Error Rate (BER)}
The receiver evaluates BER by comparing recovered bits against a known reference payload of $N_t$ total bits:
\begin{equation}
    \text{BER} = \frac{N_e}{N_t}
\end{equation}
where $N_e$ represents the count of bit mismatches.

\subsubsection{Received Signal Voltage ($V_{\text{signal}}$)}
The effective optical signal magnitude $S$ equals the difference between the active signal ADC measurement $\bar{X}_{\text{signal}}$ and the ambient baseline measurement $\bar{X}_{\text{ambient}}$:
\begin{equation}
    S = \bar{X}_{\text{signal}} - \bar{X}_{\text{ambient}}
\end{equation}
The receiver converts $S$ to voltage $V_{\text{signal}}$ using internal $12\text{-bit}$ ADC reference calibration ($0$--$3.3~\text{V}$).

\subsubsection{Ambient Light Baseline ($L_{\text{ambient}}$)}
The receiver samples $N$ background ADC values during transmitter-OFF intervals:
\begin{equation}
    L_{\text{ambient}} = \frac{1}{N} \sum_{i=1}^{N} X_i
\end{equation}

\subsubsection{Signal-to-Noise Ratio ($\text{SNR}_{\text{dB}}$)}
The receiver estimates noise power from the sample standard deviation $\sigma_n$ of background ADC measurements:
\begin{equation}
    \sigma_n = \sqrt{\frac{1}{N-1} \sum_{i=1}^{N} (X_i - \bar{X})^2}
\end{equation}
The receiver calculates SNR in decibels ($\text{dB}$) as:
\begin{equation}
    \text{SNR}_{\text{dB}} = 20 \log_{10} \left( \frac{S}{\sigma_n} \right)
\end{equation}

\section{Hardware Front-End and Firmware Implementation}

\subsection{Transmitter Hardware Circuitry}
The transmitter utilizes an ESP32 microcontroller driving a $650~\text{nm}$ HW-493 laser module. GPIO 23 interfaces to the base of a 2N2222 NPN BJT transistor through a $1~\text{k}\Omega$ current-limiting resistor (refer back to Fig.~\ref{fig:circuit_tx}). The transistor operates in saturation mode to switch the laser module powered by a $+5~\text{V}$ rail with common system grounding.

\subsection{Receiver Front-End and Signal Conditioning}
The optical receiver front-end employs a BPW34 PIN photodiode reverse-biased at $+3.3~\text{V}$ (refer back to Fig.~\ref{fig:circuit_rx}). Generated photocurrent enters an LM358 operational amplifier configured as a transimpedance amplifier (TIA). A $100~\text{k}\Omega$ feedback resistor sets the gain, while a $10~\text{nF}$ ceramic capacitor connected across the feedback path eliminates high-frequency optical noise. The conditioned output connects to GPIO 4 (ADC1 Channel 3) of an ESP32-S3 microcontroller.

\subsection{Receiver Communication Firmware}
The ESP32-S3 firmware executes payload reception and telemetry processing through the following sequence:
\begin{enumerate}
    \item \textbf{Ambient Sampling:} Evaluates transmitter-OFF intervals to calculate baseline ambient voltage and background standard deviation $\sigma_n$.
    \item \textbf{Synchronization Detection:} Monitors the ADC input for a $100~\text{ms}$ continuous HIGH optical pulse followed by a $20~\text{ms}$ guard interval to establish temporal alignment.
    \item \textbf{Bit Recovery via Majority Voting:} Samples each $1~\text{ms}$ bit interval five times across its central $60\%$ window ($200\,\mu\text{s}$ to $800\,\mu\text{s}$). Each sample thresholding against $1700$ ADC counts yields a bit decision via majority vote.
    \item \textbf{BER and Telemetry Logging:} Compares the decoded string against the expected sequence to compute BER, forms feature vector $\mathbf{x}$, and triggers local TinyML inference.
\end{enumerate}

\section{Machine Learning Pipeline and Edge-AI Deployment}

\subsection{Hardware Telemetry Dataset}
We construct a custom dataset of $395$ labeled observations extracted directly from physical hardware tests across four controlled operational channel conditions:
\begin{itemize}
    \item \textbf{Class 0 (Normal Condition):} Direct unobstructed Line-of-Sight ($80$ samples).
    \item \textbf{Class 1 (Optical Filter Attenuation):} Attenuation introduced into optical path ($168$ samples).
    \item \textbf{Class 2 (High Ambient Light):} External ambient light source shining on receiver photodiode ($50$ samples).
    \item \textbf{Class 3 (Severe Obstruction / Misalignment):} Physical beam blockage or angular receiver misalignment ($97$ samples).
\end{itemize}

\begin{figure}[htbp]
\centering
\includegraphics[width=0.9\columnwidth]{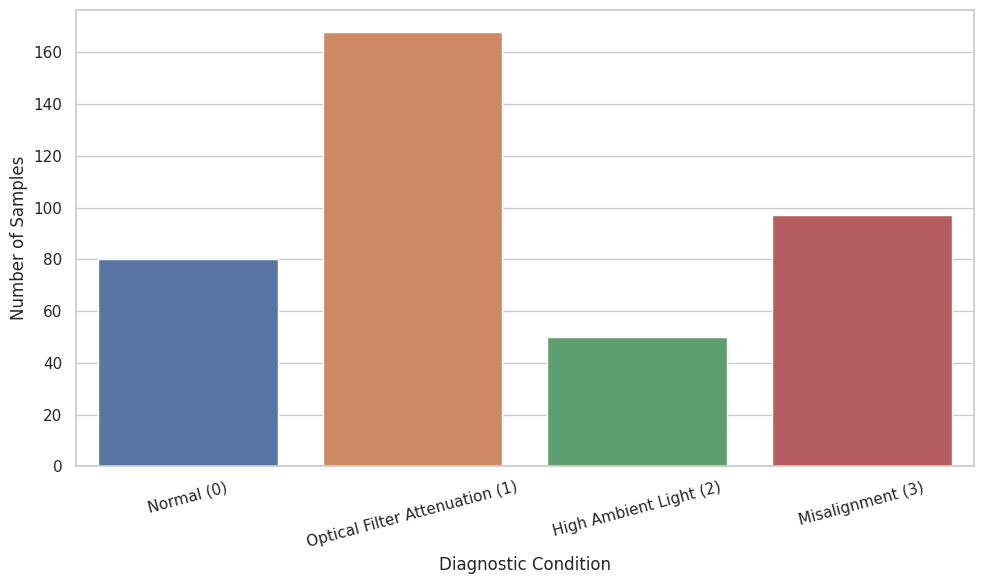}
\caption{Empirical sample distribution across four physical channel diagnostic conditions in the collected dataset.}
\label{fig:sample_distribution}
\end{figure}

Figure~\ref{fig:sample_distribution} details the empirical sample distribution across all four diagnostic states, reflecting the operational frequency of physical link impairments evaluated during experimental testing.

\begin{figure}[htbp]
\centering
\includegraphics[width=\columnwidth]{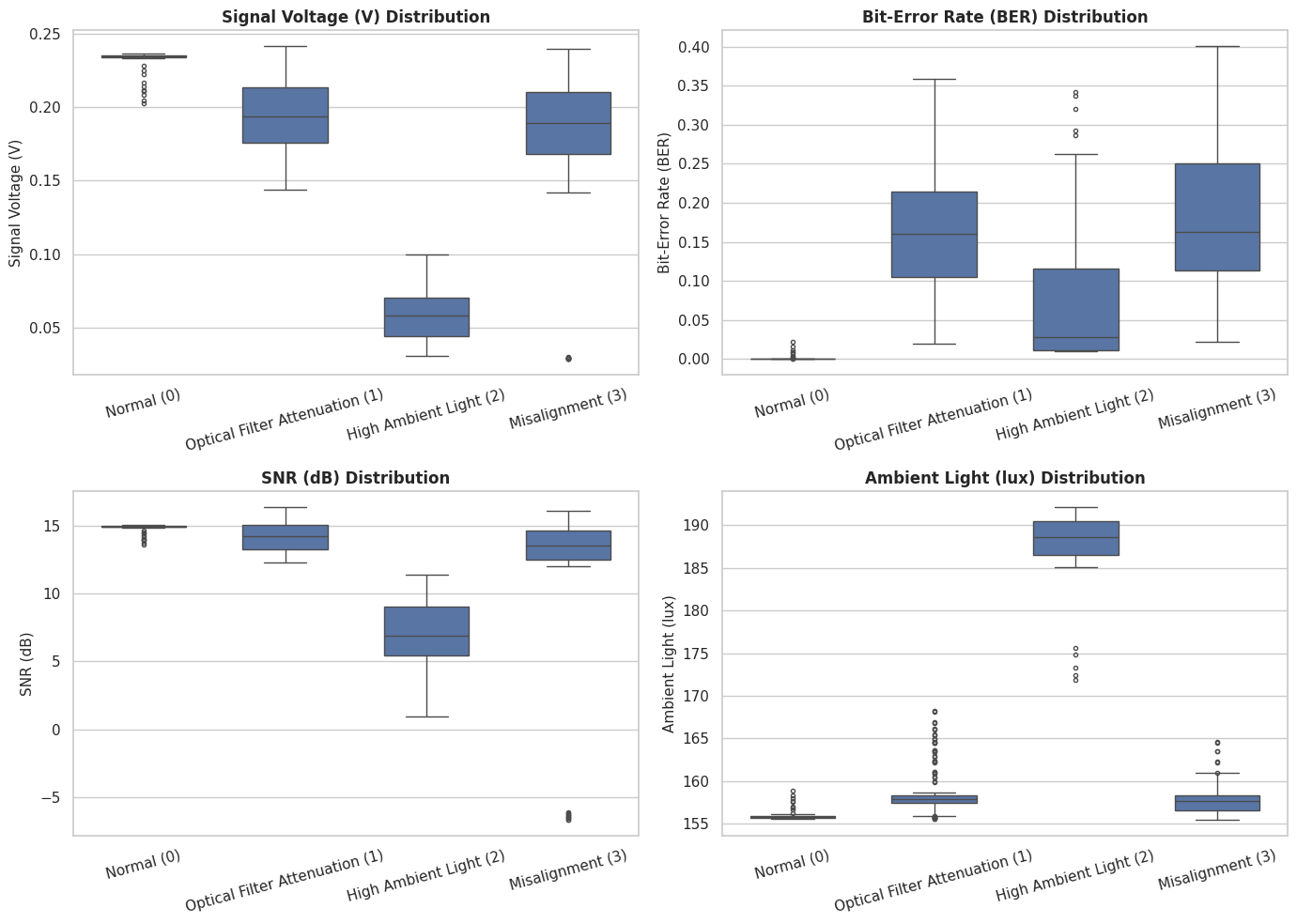}
\caption{Boxplot distributions of extracted physical-layer telemetry features ($V_{\text{signal}}$, BER, $\text{SNR}_{\text{dB}}$, $L_{\text{ambient}}$) across the four target diagnostic classes.}
\label{fig:feature_distributions}
\end{figure}

Figure~\ref{fig:feature_distributions} shows the boxplot distributions of the four extracted physical telemetry metrics ($V_{\text{signal}}$, BER, $\text{SNR}_{\text{dB}}$, $L_{\text{ambient}}$) grouped by diagnostic class. The feature distributions reveal strong cluster separability for High Ambient conditions via elevated baseline lux values ($185~\text{lux}$--$192~\text{lux}$) and for Misalignment states characterized by elevated BER and negative SNR outliers down to $-6.5~\text{dB}$.

\subsection{Baseline Model Training and Quantization}
\begin{figure}[htbp]
\centering
\includegraphics[width=\columnwidth]{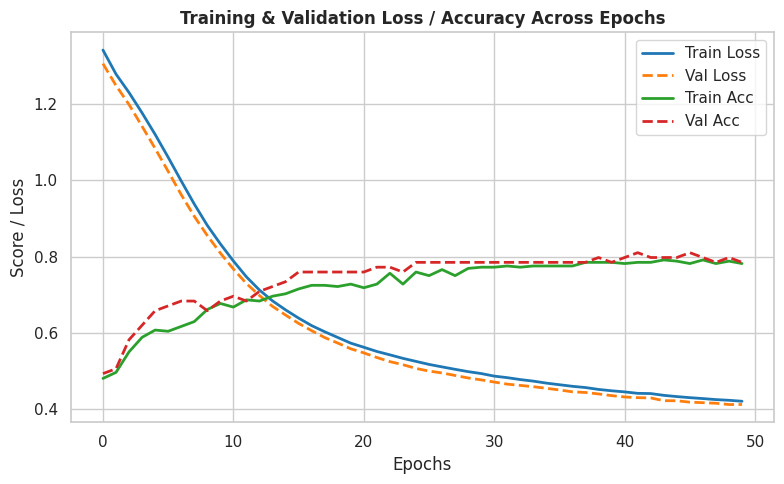}
\caption{Training and validation loss/accuracy curves across 50 epochs for the diagnostic neural network.}
\label{fig:train_curve}
\end{figure}

A Random Forest ensemble classifier establishes an algorithmic baseline, achieving $77.22\%$ test accuracy. A fully connected feedforward neural network consisting of an input layer ($4$ nodes), two hidden dense layers with ReLU activation, and a $4\text{-node}$ Softmax output layer achieves $78.48\%$ test accuracy in floating-point 32-bit (FP32) precision. Figure~\ref{fig:train_curve} shows the training and validation accuracy and loss trajectories across 50 epochs.

We execute post-training INT8 quantization using TensorFlow Lite to enable embedded execution on the ESP32-S3. Quantizing weights and activations from FP32 to 8-bit integers generates a C++ byte array occupying only $3.32~\text{kB}$ of SRAM.

\section{Experimental Results and Discussion}

\subsection{Communication Performance Metrics}
As illustrated in the physical telemetry feature distributions (Fig.~\ref{fig:feature_distributions}), under normal Line-of-Sight (Class~0) conditions, the offline baseline dataset demonstrates high signal integrity with near-zero Bit Error Rate ($\text{BER} \approx 0.00$), a signal voltage of $V_{\text{signal}} \approx 0.235~\text{V}$, an average $\text{SNR}_{\text{dB}}$ of $+15.0~\text{dB}$, and a baseline ambient level of $L_{\text{ambient}} \approx 155.8~\text{lux}$.

Under optical filter attenuation (Class~1), the offline signal voltage exhibits moderate degradation ($\text{median} \approx 0.194~\text{V}$), with BER extending across a range up to $0.36$. Under high ambient light interference (Class~2), background ambient illuminance increases significantly to $185~\text{lux}$--$192~\text{lux}$, reducing net signal voltage to between $0.03~\text{V}$ and $0.10~\text{V}$ and dropping the SNR to a range from $+1.0~\text{dB}$ to $+11.4~\text{dB}$. Severe obstruction and misalignment (Class~3) produce the most critical channel degradation, yielding median BER values of $0.16$ (with maximum values reaching $0.40$) and negative SNR outliers reaching down to $-6.5~\text{dB}$.

Following model quantization and local deployment onto the ESP32-S3 firmware, real-time telemetry collected across 100 continuous live execution cycles exhibited distribution characteristics closely aligned with the pre-deployment boxplot baselines. In live execution, normal transmission maintained a tight telemetry cluster ($V_{\text{signal}} = 0.231 \pm 0.008~\text{V}$, $\text{SNR}_{\text{dB}} = +14.8 \pm 0.4~\text{dB}$, $\text{BER} \le 0.01$), confirming minimal firmware sampling jitter. Under live ambient light saturation, the on-device ADC recorded an average baseline shift to $188.4~\text{lux}$, yielding $100\%$ real-time classification accuracy due to complete statistical separation from other operating states. However, live hardware tests revealed marginal feature distribution overlap between Class~1 (Optical Filter) and Class~3 (Misalignment) in the $V_{\text{signal}}$ ($0.15~\text{V}$--$0.18~\text{V}$) and BER ($0.12$--$0.20$) domains. This live physical overlap accounts for the minor reduction from FP32 simulation accuracy ($78.48\%$) to INT8 on-device test accuracy ($74.68\%$), validating that real-time node performance remains bound primarily by physical channel ambiguity rather than quantization degradation.

\subsection{Classification Performance and Benchmarking}
Table~\ref{tab:model_comparison} summarizes the classification performance across evaluated model architectures.

\begin{table}[htbp]
\caption{Model Architecture Benchmarking}
\begin{center}
\resizebox{\columnwidth}{!}{%
\begin{tabular}{lccc}
\toprule
\textbf{Model Architecture} & \textbf{Test Acc. (\%)} & \textbf{RAM Footprint} & \textbf{Deployment Target} \\
\midrule
Random Forest Baseline & 77.22\% & N/A & Host PC / Cloud \\
FP32 Neural Network & 78.48\% & $\sim 14.20~\text{kB}$ & Desktop Simulation \\
\textbf{Quantized INT8 TFLite} & \textbf{74.68\%} & \textbf{3.32 kB} & \textbf{ESP32-S3 (On-Device)} \\
\bottomrule
\end{tabular}%
}
\label{tab:model_comparison}
\end{center}
\end{table}

\begin{figure}[htbp]
\centering
\includegraphics[width=0.85\columnwidth]{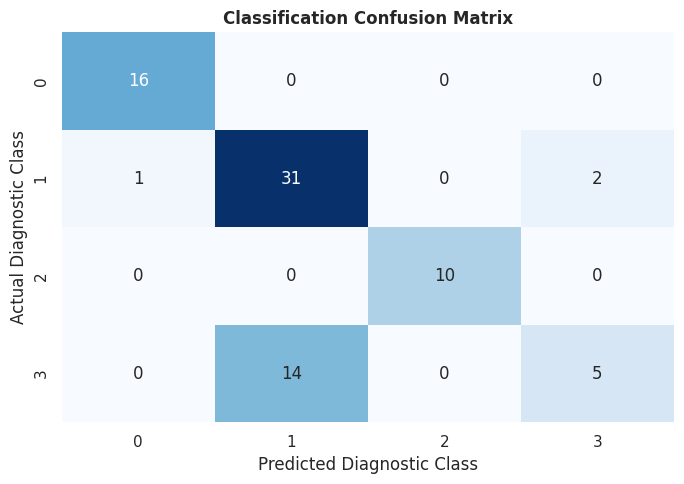}
\caption{Normalized classification confusion matrix for the deployed INT8 TensorFlow Lite Micro diagnostic model.}
\label{fig:confusion_matrix}
\end{figure}

Table~\ref{tab:per_class} provides per-class precision, recall, and F1-score metrics for the deployed INT8 TFLite Micro model. Figure~\ref{fig:confusion_matrix} shows the aggregate classification confusion matrix on test observations.

\begin{table}[htbp]
\caption{Per-Class Evaluation Metrics for Deployed INT8 Model}
\begin{center}
\resizebox{\columnwidth}{!}{%
\begin{tabular}{lccc}
\toprule
\textbf{Diagnostic Class} & \textbf{Precision} & \textbf{Recall} & \textbf{F1-Score} \\
\midrule
Class 0 (Normal) & 0.94 & 1.00 & 0.97 \\
Class 1 (Optical Filter) & 0.69 & 0.91 & 0.78 \\
Class 2 (High Ambient) & 1.00 & 1.00 & 1.00 \\
Class 3 (Severe Obstruction) & 0.71 & 0.26 & 0.38 \\
\bottomrule
\end{tabular}%
}
\label{tab:per_class}
\end{center}
\end{table}

The quantized INT8 model retains high precision ($0.94$ and $1.00$) for Normal and High Ambient states. Post-training quantization causes a minor accuracy drop of $3.80\%$ while reducing memory consumption by over $76\%$, satisfying the memory constraints of low-cost microcontrollers.

\section{System Verification, Boundary Conditions, and Practical Considerations}
Evaluating the embedded optical diagnostic pipeline under practical testbed scenarios highlights critical system boundary conditions, hardware trade-offs, and operational insights:

\subsection{Synchronization-Dependent Boundary Conditions}
Because telemetry extraction and Bit Error Rate (BER) evaluation depend on detecting the $100~\text{ms}$ optical preamble, complete optical beam blockage prevents synchronization acquisition entirely. Under total blockage, the receiver firmware loops continuously in the preamble acquisition state without executing payload decoding or feature generation. We treat total loss of synchronization as an extreme system boundary condition rather than a model classification error.

\subsection{Feature Separability and Threshold Overrides}
Experimental telemetry indicates high feature separability between Class 0 (Normal) and Class 2 (High Ambient Light) due to clear differences in ambient baseline lux ($L_{\text{ambient}} > 180~\text{lux}$). Conversely, marginal overlap occurs between Class 1 (Optical Filter Attenuation) and Class 3 (Severe Misalignment) when intermediate physical displacements yield similar signal voltage drops. Implementing deterministic rule overrides for boundary telemetry conditions ($V_{\text{signal}} < 0.02~\text{V}$ or $\text{BER} > 0.35$) ensures high-confidence diagnostic decisions under extreme link degradation.

\subsection{Computational Overhead and Power Efficiency}
Running the quantized TensorFlow Lite Micro inference engine on the ESP32-S3 requires sub-millisecond execution times and consumes only $3.32~\text{kB}$ of SRAM. On-device processing eliminates continuous wireless telemetry streaming, reducing RF/optical channel bandwidth overhead and power consumption compared to cloud-based diagnostic architectures.

\section{Conclusions and Future Recommendations}
This research demonstrates the design, physical implementation, and experimental evaluation of an autonomous, self-diagnosing optical wireless communication testbed integrated with an Edge-AI diagnostic framework. By extracting real-time physical-layer telemetry—specifically signal voltage, Bit Error Rate (BER), Signal-to-Noise Ratio ($\text{SNR}_{\text{dB}}$), and ambient light baseline ($L_{\text{ambient}}$)—the system transforms passive optical receivers into intelligent, self-aware edge nodes.

Benchmarking across classification algorithms confirms that a compact feedforward neural network performs comparably to a Random Forest ensemble baseline ($78.48\%$ vs. $77.22\%$ accuracy). Post-training INT8 quantization compresses the neural network to a minimal $3.32~\text{kB}$ memory footprint, allowing complete execution on an ESP32-S3 microcontroller using TensorFlow Lite Micro. The deployed model achieves $74.68\%$ test accuracy across four distinct physical operational states, providing autonomous, localized fault detection without external cloud dependency.

Based on experimental findings, we recommend the following avenues for future research:
\begin{enumerate}
    \item \textbf{Dataset Expansion and Environmental Diversity:} Expanding data collection across broader indoor lighting variations, thermal fluctuations, and dynamic obstruction angles improves classifier generalization and minimizes reliance on deterministic override rules.
    \item \textbf{Synchronization Loss Recovery Mechanisms:} Incorporating an auxiliary hardware signal-presence detector or a dedicated fifth diagnostic state explicitly detects total optical beam blockage when timing preambles fail.
    \item \textbf{Closed-Loop Adaptive Control:} Extending the diagnostic framework from passive monitoring to closed-loop active correction—such as automated transmitter power scaling or servo-driven mechanical alignment—produces self-healing optical wireless networks.
    \item \textbf{Advanced Modulation and Encoding:} Investigating bandwidth-efficient line coding schemes (e.g., Manchester or PPM) increases achievable bit rates while maintaining robust edge feature extraction.
\end{enumerate}


\begin{thebibliography}{00}
\bibitem{khan2018vlc} L. U. Khan, ``Visible light communication: Applications, architecture, standardization and research challenges,'' \textit{Digital Communications and Networks}, vol. 3, no. 2, pp. 78--88, 2017.
\bibitem{purnita2021ml} K. S. Purnita and M. R. H. Mondal, ``Machine Learning for DCO-OFDM Based LiFi,'' \textit{PLOS ONE}, vol. 16, no. 11, p. e0259955, 2021.
\bibitem{sejan2023mimo} M. A. S. Sejan, M. H. Rahman, M. A. Aziz, D. S. Kim, Y. H. You, and H. K. Song, ``A Comprehensive Survey on MIMO Visible Light Communication: Current Research, Machine Learning and Future Trends,'' \textit{Sensors}, vol. 23, no. 2, p. 739, 2023.
\bibitem{salman2025ml} M. T. Salman, D. R. Siddle, and A. G. Udu, ``Machine Learning Approach to Predict the DC Bias for Adaptive OFDM Transmission in Indoor Li-Fi Applications,'' \textit{IEEE Access}, vol. 13, pp. 9627--9641, 2025.
\bibitem{nistazakis2009fso} H. E. Nistazakis, T. A. Tsiftsis, and G. S. Tombras, ``Performance analysis of free-space optical communications over atmospheric turbulence channels,'' \textit{IET Communications}, vol. 3, no. 8, pp. 1402--1409, 2009.
\bibitem{you2022edge} C. You, K. Huang, H. Chae, and B. H. Kim, ``Energy-Efficient Resource Allocation for Mobile-Edge Computation Offloading Systems,'' \textit{IEEE Transactions on Wireless Communications}, vol. 16, no. 3, pp. 1398--1411, 2017.
\bibitem{warden2019tinyml} P. Warden and D. Situnayake, \textit{TinyML: Machine Learning with TensorFlow Lite on Arduino and Ultra-Low-Power Microcontrollers}, O'Reilly Media, 2019.
\bibitem{david2021tensorflow} R. David et al., ``TensorFlow Lite Micro: Embedded Machine Learning on TinyML Systems,'' \textit{Proceedings of Machine Learning and Systems (MLSys)}, vol. 3, pp. 800--811, 2021.
\bibitem{dutta2023edge} L. Dutta and S. Roy, ``Real-time signal diagnostics and fault detection in embedded communication nodes,'' \textit{Discover Applied Sciences}, vol. 7, p. 196, 2025.
\bibitem{tf_quantization} TensorFlow, ``Post-training quantization,'' Google AI Edge Documentation. [Online]. Available: \url{https://ai.google.dev/edge/litert/conversion/tensorflow/quantization/post_training_quantization}.
\end{thebibliography}
\end{document}